\documentclass[10pt,twocolumn,letterpaper]{article}

\usepackage{3dv}
\usepackage{times}
\usepackage{epsfig}
\usepackage{graphicx}
\usepackage{caption}
\usepackage{subcaption}
\usepackage{amsmath}
\usepackage{mathtools} 
\usepackage{amssymb}
\usepackage{algorithm} 
\graphicspath{{images/}}
\usepackage{algpseudocode}
\usepackage{amsthm}
\usepackage[labelfont=bf]{caption}
\newcommand\norm[1]{\left\lVert#1\right\rVert}

\usepackage[pagebackref=true,breaklinks=true,colorlinks,bookmarks=false]{hyperref}
\usepackage{cleveref}

\usepackage{enumitem}
\setlist{topsep=0pt, leftmargin=*,noitemsep,topsep=0pt,parsep=0pt,partopsep=0pt}

\threedvfinalcopy 

\usepackage{cuted}
\usepackage{capt-of}

\begin{document}

\title{Convex Joint Graph Matching and Clustering via Semidefinite Relaxations}

\author{ 
\begin{tabular}{ccc} 
Maximilian Krahn$^{1,2,3}$ & Florian Bernard$^{4,5}$ & Vladislav Golyanik$^1$ 
\end{tabular} 
\vspace{11pt}
\\ 
\normalsize
\begin{tabular}{ccccc} 
$^1$MPI for Informatics, SIC & $^2$Saarland University & $^3$Aalto University & $^4$University of Bonn & $^5$TU Munich 
\end{tabular} 
} 

\maketitle

\begin{abstract} 
This paper proposes a new algorithm for simultaneous graph matching and clustering. 
For the first time in the literature, these two problems are solved jointly and synergetically without relying on any training data, which brings advantages for identifying similar arbitrary objects in compound 3D scenes and matching them. 
For joint reasoning, we first rephrase graph matching as a rigid point set registration problem operating on spectral graph embeddings. 
Consequently, we utilise efficient convex semidefinite program relaxations for aligning points in Hilbert spaces and add coupling constraints to model the mutual dependency and exploit synergies between both tasks. 
We outperform state of the art in challenging cases with non-perfectly matching and noisy graphs, and we show successful applications on real compound scenes with multiple 3D elements. 
Our source code and data are publicly available. 
\end{abstract}

\section{Introduction} 
Graph matching and graph clustering are fundamental and long-standing computer vision problems that are typically treated separately.
The most common way to address 
both problems simultaneously is by applying 
two different methods to the same input graphs, which, unfortunately, does not always lead to desired and accurate results. 
This often happens because the input data is not segmented in advance, and such segmentation can be difficult to obtain  without additional prior knowledge. 
On the other hand, general-purpose clustering methods such as $k$-means or general-purpose matching methods, are not aware of the compound nature of the data, and perform sub-optimally. 

\begin{figure}[t!]
    \centering
    \includegraphics[width = \linewidth]{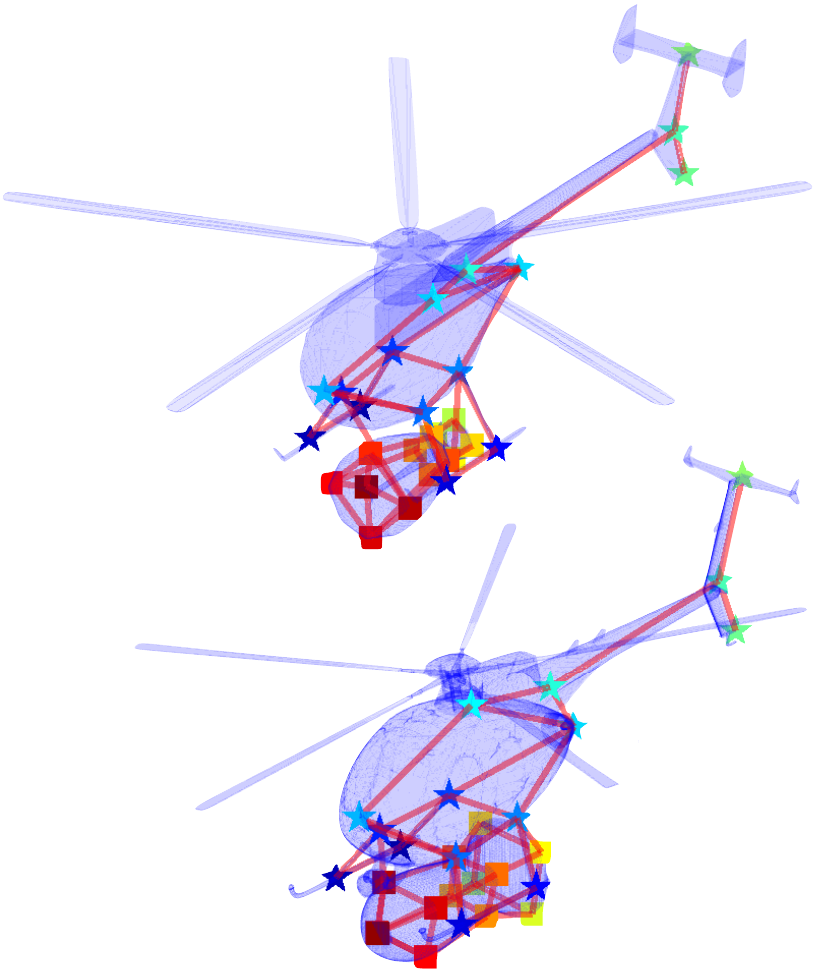}
    \caption{Our method for joint graph matching and clustering obtains correspondences (node colour) and segments of challenging graphs (node marker), as shown in this example from the Princeton shape dataset \cite{princtonShape}. 
    Note that vases carried by the helicopters are cylindrically symmetric. 
    } 
    \label{fig:teaserHeli}
\end{figure}

\begin{figure}\centering
\includegraphics[width=\linewidth]{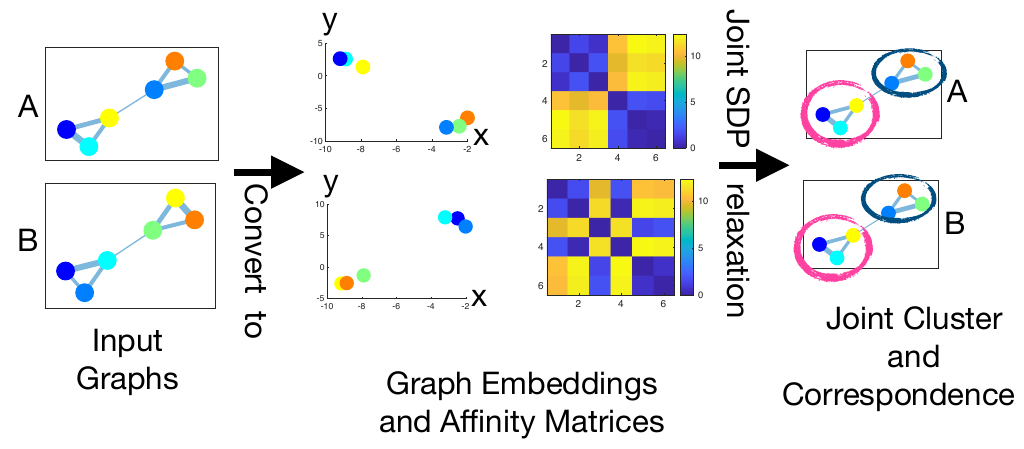}
\captionof{figure}{\textbf{An overview of the proposed algorithm for joint graph matching and clustering.} 
The inputs are two graphs (left), which are embedded as point clouds with the corresponding affinity matrices (middle). 
Next, we solve our joint formulation with a novel SDP  relaxation and obtain the clustering (coloured circles)  and graph matching (\textit{i.e.,} every matched node has the same colour) as a result. 
\label{fig:feature-graphic}} 
\end{figure}

Motivated by these observations, we investigate and propose in this paper the first non-learning-based method for simultaneous graph matching and clustering with mutual constraints, see Fig.~\ref{fig:teaserHeli} for an exemplary solution and Fig.~\ref{fig:feature-graphic} for an overview. 
\textit{Despite being an important problem and a promising direction to explore, joint graph matching and clustering has not been thoroughly researched yet.} 
We observe in experiments that graph matching can improve the accuracy of clustering in challenging scenarios, and vice versa, when both problems are considered jointly. 
The new method shows improved characteristics compared to the current state of the art, thanks to the new form of our joint convex semi-definite program (SDP) relaxation, with its solution optimality guarantee. 
To this end, we add novel coupling constraints to SDP relaxations of the graph matching and clustering problems, which introduces the mutual dependency between the estimates.
In other words, the solutions to both problems serve as additional mutual cues, either for clustering or for matching. 
Our approach tackles challenging scenarios and resolves cases that can be ambiguous for disjoint techniques. 
To summarise, the primary technical \textbf{contributions} of this  paper are as follows: 
\begin{enumerate} 
    \item A novel approach for graph matching and clustering by a joint SDP formulation. Our technique increases the accuracy of both matching and clustering in challenging scenarios compared to solutions obtained by methods solving these problems separately. 
    \item A novel formulation of graph matching with  the problem interpreted as point set alignment of graph embeddings. Its convex SDP relaxation leads to  increased efficiency, and we achieve an  order-of-magnitude faster matching compared to the graph matching
    method of Schellewald \textit{et al.}~\cite{schellewald} using convex SDP relaxation. 
\end{enumerate} 

Experimentally, we evaluate how accurately our method matches graphs on the  CMU House dataset \cite{CMUHouse}. 
To measure the joint matching and clustering performance, we compare our algorithm against other graph matching and clustering algorithms on synthetic data. 
Moreover, we demonstrate the effectiveness of joint clustering and  matching on selected and customised samples of the \textit{Princeton Shape} dataset \cite{princtonShape}. 
The source code and the data are available for research  purposes\footnote{\url{https://4dqv.mpi-inf.mpg.de/JointGMC/}}. 

\section{Background and Related Work} 
This section introduces notations and problem formulations (Sec.~\ref{ssec:problem_formulations}) and reviews related works (Sec.~\ref{ssec:related_works}). 
\subsection{Notations and Problem  Definitions}\label{ssec:problem_formulations} 
\begin{figure}
    \centering
    \includegraphics[width = \linewidth]{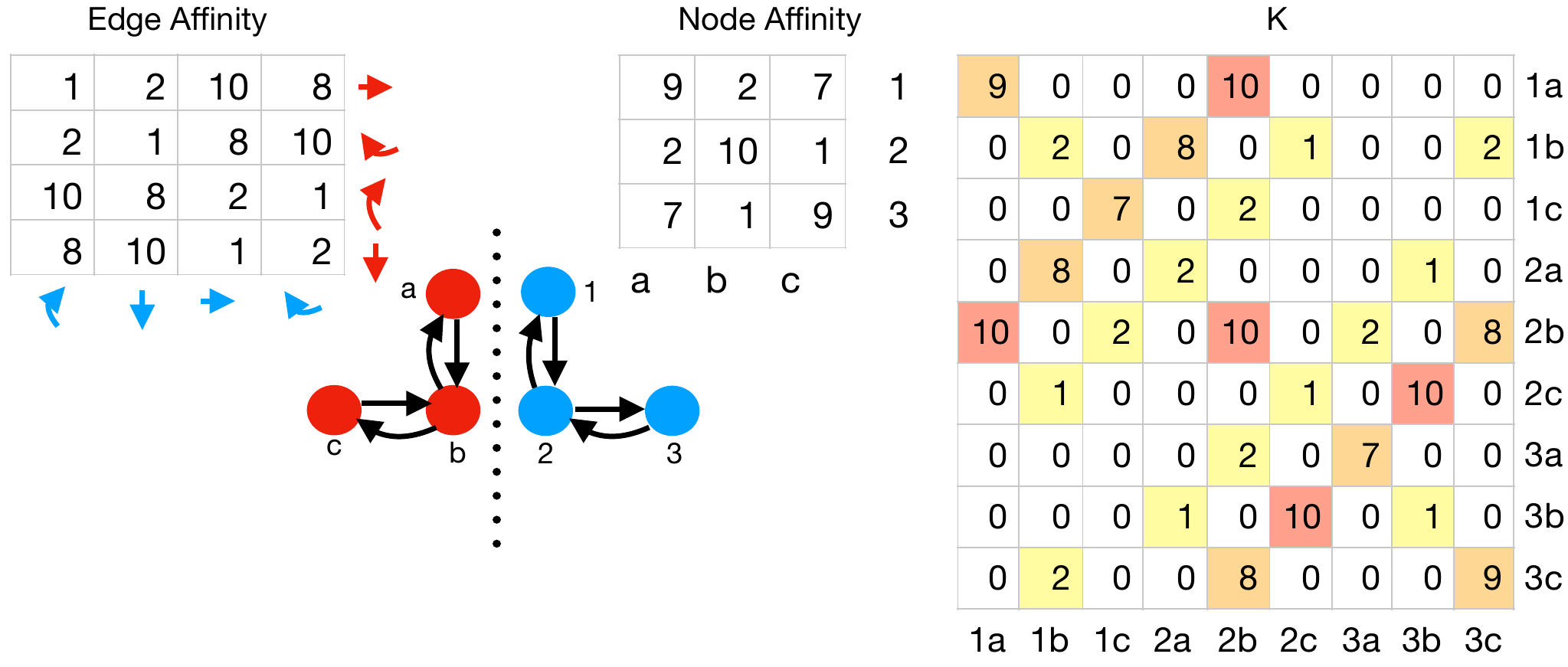}
    \caption{A visualisation of the edge affinity, node affinity and the affinity matrix $K$ for an exemplary synthetic graph matching problem at the bottom-left. 
    }
    \label{fig:affinityMatrix}
\end{figure}

A graph $\mathcal{G} = (\mathcal{V}, \mathcal{E})$ consists of nodes (vertices) $\mathcal{V}$ and edges $\mathcal{E}$, each connecting two nodes. 
We assume in this paper that $\mathcal{E}$ are weighted, and $\mathcal{G}$ can be directed. 
A useful graph representation is an adjacency matrix denoted by $A$, which shows  which nodes are connected and which weights the edges have. 
As a similarity measure between two graphs $\mathcal{G}_1$ and  $\mathcal{G}_2$, we use an affinity matrix $K$, see  Fig.~\ref{fig:affinityMatrix}. 
$K$ is constructed according to the following rules \cite{fgm}: 
1) An entry $K_{i_1 i_2,j_1j_2 }$ is equal to the node affinity of $i_1$  and $i_2$, if and only if $ i_1 = j_1$ and $ i_2 = j_2$; 
2) Otherwise, $K_{i_1 i_2,j_1j_2 }$ is equal to the edge affinity of $i_1  j_1$ and $i_2 j_2$. 
Here, $\{i_1$, $j_1\}$ denote node pairs of $\mathcal{G}_1$ and $\{i_2$,  $j_2\}$ denote node pairs of $\mathcal{G}_2$. 
If an edge or a node does not exist, $K_{i_1 i_2,j_1j_2 } = 0$. 

\textbf{Graph Matching.} 
The goal of graph matching is to find correspondences between the nodes of two graphs  $\mathcal{G}_1$ and $\mathcal{G}_2$, while respecting respective edge weights. 
The historically first formulation of the graph matching (GM)  problem is known as the quadratic assignment problem (QAP) in Koopmans-Beckmann form  \cite{Beckman1957}. 
Given two adjacency matrices $A$ and $B$ of two graphs, the goal is to find a permutation $\Pi_n \coloneqq \{X \in  \{0, 1\}^{n \times n} : X \boldsymbol{1}_n = \boldsymbol{1}_n, \boldsymbol{1}_n^\mathsf{T} X = \boldsymbol{1}_n^\mathsf{T} \}$ which maximises a quadratic objective, \textit{i.e.,} 
\begin{equation} 
  \underset{X \in \Pi_n}{\max} \operatorname{tr}(A  X^\mathsf{T} B X). 
    \label{eq:Koopmans-Beckmann} 
\end{equation} 
The Lawler form is a more general formulation of the quadratic assignment problem \cite{Lawler1963}. 
It is defined using the  inter-graph affinity matrix  $K$ and reads 
\begin{equation}
  \underset{X \in \Pi_n}{\max} \operatorname{vec}\left(X  \right)^T K \operatorname{vec}\left( X \right). 
    \label{eq:lawler} 
\end{equation} 
Every Koopmans-Beckmann form can be converted to the Lawler form by  constructing 
$K := A \otimes B$. 

\begin{figure}
    \centering
    \includegraphics[width = 0.3\textwidth]{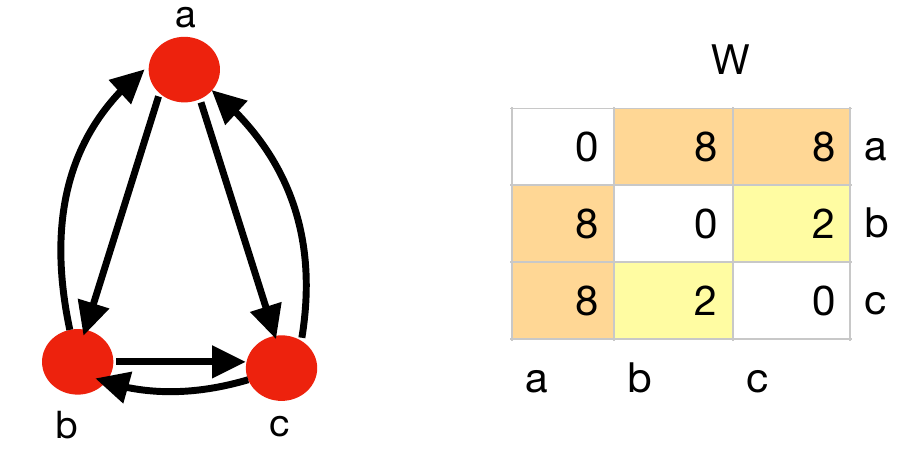}
    \caption{Visualisation of an intra-graph affinity  matrix $W$ used for clustering. 
    \textit{Left:} An exemplary synthetic graph; 
    \textit{right:} an affinity matrix based on its node coordinates.} 
    \label{fig:WAffinity} 
\end{figure}

\textbf{Graph Clustering.} 
The goal of graph clustering is to segment the input  $\mathcal{G}$ into meaningful sub-graphs. 
Similarly to the inter-graph affinity $K$, we also define intra-graph affinity $W$ visualised in Fig.~\ref{fig:WAffinity}. 
Each entry $W_{ij}$ is defined as the affinity score between two nodes $i$ and $j$ of $\mathcal{G}$. 
We map the nodes to Euclidean coordinates; those are  either provided from the data or calculated with  embeddings. 
After that, we define each entry $W_{ij}$ as the Euclidean  distance between those nodes. 
This score can be taken as a penaliser, \textit{i.e.,} the higher the score, the more dissimilar are the two  nodes. 
Consequently, we can use the maximum cut (MAX CUT)  formulation, like done by Goemans and Williamson  \cite{maxcutSDP}, to define the clustering problem for two segments as the optimisation problem 
\begin{equation}
     \underset{y \in \{ -1,1\}^n}{\max} \sum_i \sum_j W_{ij}\,(1-y_iy_j)\,, 
\label{eq:MAXCUT}
\end{equation}
where  $y$ defines the cluster assignment of the nodes. 

\subsection{Related  Work}\label{ssec:related_works} 
\textbf{Graph Matching.} 
Both formulations of graph matching  \eqref{eq:Koopmans-Beckmann} and \eqref{eq:lawler} are $\mathcal{NP}$-hard, and no algorithms are known that can solve general graph matching instances in polynomial time. 
Hence, in addition to branch and bound approaches that have exponential worst-case time complexity~\cite{hahn1998branch}, most works focus on finding approximate solutions. There is an increasing interest in theoretically well-grounded solutions for graph matching. Such methods are, for example, based on dual decomposition~\cite{Torresani2013,Swoboda2017b}, or rely on spectral~\cite{firstSpectral,sm,Cour2006,Egozi2012} or convex relaxations~\cite{schellewald}, often in combination with a convex-to-concave path-following approach in order to find feasible solutions to the non-relaxed problem~\cite{Zaslavskiy2009,fgm,Dym2017,bernard2018dsstar}. 

Other graph matching methods utilise integer-projected fixed point methods~\cite{Leordeanu2009}, random walks~\cite{cho2010RRWM}, max-pooling~\cite{Cho2014}, gradient-descent type optimisation~\cite{koopmanBeckmanRecentAlg}, belief propagation~\cite{Zhang2016}, or the alternating direction method of multipliers~\cite{LeHuu2017}. 
Since the QAP constitutes one popular way of phrasing the graph matching problem, algorithms that
 target the QAP  can readily be used for graph matching ~\cite{Pardalos1994, Zhao1998, loiola2007survey,Olsson2007, kezurer2015,admm}. The recent quantum method~\cite{SeelbachBenkner2020} attempts solving graph matching without  relaxation and can at the moment only be used for small problem instances. 

Learning-based graph matching becomes an increasingly active  research direction \cite{cho2013HARG, zhang2013Oriented},  reinforced by recent advances in deep learning~\cite{zanfir2018gmn, wang2019PCA, rolinek2020deep, fey2020deep}. 
The downside of these approaches is that they often rely on collections of labelled data, which might not always be available. 
In this work, we instead focus on an approach which does not have such strong dependencies. 

\textbf{Graph Clustering.} 
Graph clustering can be used for grouping information provided as graphs. 
A common approach is to embed graph nodes with a graph embedding into a Hilbert space  and subsequently perform clustering (\textit{e.g.,} using $k$-means)  \cite{spectralEmbedding, HOPE, cao2015grarep}. 
However, instead of reducing the dimension of the adjacency matrix, we can also use MAX CUT \eqref{eq:MAXCUT} \cite{hartmanis1982computers} to find two graph clusters directly.
The advantage of this formulation is that we do not rely on the quality of the graph embedding. 
However, it was proven that this problem is $\mathcal{NP}$-hard and, therefore, not efficiently solvable in polynomial time. 

Hence, most works in recent years focused on finding more efficient approximation algorithms. 
One way to approximate MAX CUT is by randomised algorithms, which was done by De La Vega \textit{et al.}~\cite{de2001randomized} for the metric MAX CUT problem; their method runs in polynomial time and finds arbitrarily close approximations. 
A different heuristic strategy is presented  in work by Festa and colleagues \cite{festa2002randomized}. 
Similarly to graph matching, spectral approaches have been considered for the MAX CUT problem  \cite{poljak1995solving, trevisan2012max}. 
Moreover, semidefinite programming (SDP) solutions have also been considered.
The popular Goemans-Williamson SDP relaxation \cite{maxcutSDP} was  shown to solve the MAX CUT problem nearly optimally in linear time given information about the graph like node degrees \cite{arora2007combinatorial}. 
Also, intersections between semidefinite and polyhedral relaxations have been discovered in \cite{rendl2010solving}. 
In addition to the MAX CUT formalism, there exist alternative formulations for graph clustering (or grouping), among them being a MIN CUT formulation~\cite{johnson1993min}, normalised cuts \cite{xu2009fast}, and multicuts~\cite{kappes2016multicuts}. 

\textbf{Solving Graph Matching and Clustering Together.} 
Joint graph matching and graph clustering is not widely researched and there  exists currently no non-learning-based algorithm that solves the problem in a truly joint manner. 
One of the first works that considered graph clustering and graph matching  together is \cite{SeededGM}. 
Here, the two graphs are clustered first, and subsequently, a  graph matching is performed within the clusters. As such, both problems are solved consecutively, rather than in a joint manner. 
A recent paper that considers graph matching and graph clustering is  \cite{multiGraphCluster}.
However, it merely relies on the standard graph matching solver RRWM~\cite{RRWM} in order to introduce clustering constraints. While this improves the accuracy of multi-graph matching, it is not a truly joint formulation of graph matching and clustering. 
A recent learning-based approach for graph matching and graph clustering can be found in \cite{wang2020graduated}. 
While this method comes close to our approach in terms of the joint problem definition, their formulation considers the matching and clustering separately in an alternating manner. 
Moreover, the approach is embedded in a learning-based method, whereas ours does not consider learning and, thus, does not require labelled training data. 
In \cite{yu2018joint},
a graph cut and matching problem is solved simultaneously.
However, in this work the aim is to cut a single graph into two clusters, and then perform a matching between these clusters.
In contrast, we use two graphs as input, and our aim is to identify clusters in each graph, while---at the same time---matching nodes between both graphs in such a way that the matching is consistent with the clustering (\textit{i.e.,} all nodes within a cluster must be matched to nodes that all lie in the same cluster of the other graph). 

Overall, while some works consider related but different settings of joint graph matching and clustering, our work proposes the first solution that  solves inter-graph matching and clustering in a truly joint manner. 

\section{Method}\label{secalgorithm} 
In this section we derive our proposed algorithm for joint graph matching and clustering. 
Our core idea is to introduce coupling constraints between graph matching and graph clustering.
Doing so allows exploiting synergies in solving both problems simultaneously, while being robust to noise. 
Moreover, our method consists of a sequence of analytically derived  steps and does not depend on the availability of training data. 
The method's derivation is split into two parts: 
\textit{First}, we present the new objective for joint graph clustering and graph matching in Sec.~\ref{ssec:joint_formulation}. \textit{Second}, we introduce a 
convex semidefinite programming (SDP) relaxation of the proposed objective in Sec.~\ref{ssec:SDP_relaxation}. 
\subsection{Joint Graph Matching and  Clustering}\label{ssec:joint_formulation} 
In our formulation, we focus on bijective matchings, where each graph has $n$ nodes. In the case of non-bijective problems, one can introduce dummy nodes.
We use the general Lawler form of the QAP for matching two graphs $\mathcal{G}_1$ and $\mathcal{G}_2$ \eqref{eq:lawler}. 
Let $K$ be our affinity matrix of size $n^2 \times n^2$, and $x = \operatorname{vec}(X)$ be a short-hand notation for the vectorised permutation matrix $X \in \Pi_n$ with stacked columns. 
With that, the graph matching problem can be written as
\begin{equation}\label{eq:GM_Lawler} 
        \underset{X \in \Pi_n}{\text{max }} x^T K x.
\end{equation} 
For graph clustering, we consider the MAX CUT formulation for two graphs at the same time, which reads 
\begin{align}
       &\underset{ y^{(1)},  y^{(2)}}{\text{max }} &&\sum_{i,j} W^{(1)}_{ij}(1{-}y^{(1)}_iy^{(1)}_j) + \sum_{i,j} W^{(2)}_{ij}(1{-} y^{(2)}_i  y^{(2)}_j),
\end{align}
where $y^{(1)}, y^{(2)} \in \{-1,1\}$ and $ W^{(1)} = [ W^{(1)}_{ij}]_{ij}$ and $ W^{(2)} = [ W^{(2)}_{ij}]_{ij}$ are  intra-graph  affinity matrices of size $n \times n$ for the graphs $\mathcal{G}_1$ and $\mathcal{G}_2$, respectively. 

Next, we introduce a  constraint that  couples the above graph matching and clustering formulations. The main  motivation is that a node $i$ of graph $\mathcal{G}_1$ can only be matched to a node $j$ of graph $\mathcal{G}_2$, if (and only if) both nodes belong to the same cluster. We express this as the (non-convex) quadratic inequality
\begin{equation}\label{eq:coupling_equation} 
    X_{ij} \leq \frac{ y^{(1)}_{i} y^{(2)}_{j} + 1}{2}. 
\end{equation}
If $X_{ij} = 0$, the constraint is always satisfied, independent from the respective cluster assignment. If $X_{ij} = 1$, the right-hand side can only satisfy the constraint if $i$ and $j$ belong to the same cluster.
Overall, our (non-convex) \textbf{joint graph matching and clustering problem} is formulated as
\begin{align}\label{eq:joint}
       &\underset{X \in \Pi_n, y^{(1)},  y^{(2)}}{\text{max }} &&x^TKx + \sum_{i,j} W^{(1)}_{ij}(1{-}y^{(1)}_iy^{(1)}_j) \\
       &&& \qquad\quad+ \sum_{i,j} W^{(2)}_{ij}(1{-} y^{(2)}_i  y^{(2)}_j) \nonumber\\ 
     &~~~\text{s.t.} &&y^{(1)}, y^{(2)} \in \{ -1,1\}^n, X_{ij} \leq \frac{ y^{(1)}_{i} y^{(2)}_{j} {+} 1}{2}.\nonumber
\end{align}

\subsection{Joint SDP Relaxation}\label{ssec:SDP_relaxation} 
In the following, we introduce our efficient convex semidefinite programming (SDP) relaxation of Problem~\eqref{eq:joint}. To this end, we first present the SDP relaxation for the  graph matching part, followed by the SDP relaxation of the clustering part, and eventually we present the SDP relaxation of the coupling constraints. 
Note that as this is a convex relaxation, we can obtain a \emph{globally optimal} solution of the relaxed problem. 
\textbf{SDP Relaxation for Graph Matching.} 
It is well-known that common graph matching SDP relaxations (as \textit{e.g.,} considered in~\cite{schellewald}) are inefficient due to the large SDP constraints of size $\mathcal{O}(n^2 {\times} n^2)$. 
To avoid this limitation, we consider a low-dimensional embedding of graph adjacency matrices instead, so that the graph matching problem is transformed to a point cloud registration problem, for which more efficient SDP relaxations are available. 

Since we consider Lawler's form of the QAP, we first convert the affinity matrix $K$ of size $n^2 \times n^2$ into multiple adjacency matrices $n \times n$ based on the Kronecker-Produkt singular value decomposition (KPSVD) method by van Loan and Pitsianis~\cite{KPSVD}. This amounts to factorising 
\begin{align}\label{eq:K_factoriation} 
K = \sum_{i =1}^{n^2} A_i \otimes B_i
\end{align} 
in terms of the adjacency matrices $\{A_i \}$ and $\{B_i\}$.
Among all matrices $A_i$ and $B_i$, we select $k \leq n^2$ that correspond to the $k$ most dominant singular values (\textit{cf.}~\cite{KPSVD}). 
Defining the index set of the considered matrices as $I = \{ 1, \dots , k \}$,  we can thus approximate the Lawler's QAP by summing over multiple Koopmans-Beckmann QAPs with matrices $A_i$ and $B_i$~\cite{fgm}, \textit{i.e.,}~
\begin{align} 
     \underset{X \in \Pi_n}{\max} \sum_{i \in I}  \operatorname{tr}\left( A_i X B_i X^\mathsf{T}\right). 
\end{align} 
Subsequently, for each of these adjacency matrices we compute a low-dimensional embedding, which allows to consider a point cloud registration problem in the embedded space~\cite{bai2004graph}.
To this end, we use the HOPE \cite{HOPE} embedding with the common neighbour index, which we found to be the most noise-resistant embedding among more than ten graph embeddings that we evaluated.
By denoting the embeddings of $A_i$ and $B_i$ 
as $P^{(1)}_{i}, P^{(2)}_{i}$ and $Q^{(1)}_{i}, Q^{(2)}_{i}$, respectively, graph matching expressed as (orthogonal) point cloud registration problem in the embedded space reads 
\begin{equation}\label{eq:GM_relaxation} 
\small 
\begin{aligned} 
     \underset{\substack{\{R^{(1)}_{i} \in O(d)\}\\ \{R^{(2)}_{i} \in O(d)\}\\ X \in \Pi_n}}{\text{min}}\sum_i  \norm{ R^{(1)}_{i} P^{(1)}_{i} {-} Q^{(1)}_{i} X }^2 {+} 
     \norm{ R^{(2)}_{i} P^{(2)}_{i} {-} Q^{(2)}_{i} X }^2. 
\end{aligned} 
\end{equation} 
Here, in addition to the permutation matrix $X$ that brings the points into correspondence, we optimise for orthogonal transformations $R_{i,1}$ and $R_{i,2}$ that spatially align the $d$-dimensional embeddings. 
The intuition behind the relation between point cloud registration and graph matching is that graph adjacency matrices can be factorised in a symmetric manner; the graph matching then leads to a point cloud registration problem in the embedded space. 

To obtain an efficient convex relaxation of Problem~\eqref{eq:GM_relaxation}, we follow the strategy proposed by Maron \textit{et al.}~\cite{PMSDP}. The main difference to their approach is that in our case we sum over multiple terms in the objective function, rather than a single one. The strong advantage of the approach~\cite{PMSDP} is that it reduces SDP constraints of size $\mathcal{O}((d+n)^2 {\times} (d+n)^2)$ to multiple smaller SDP constraints of size $\mathcal{O}((d^2+n) {\times} (d^2+n))$, where $d$ is the dimensionality of the embedding. For more details, we refer readers to~\cite{PMSDP}.

\textbf{SDP Relaxation for Graph Clustering.} 
To obtain a convex relaxation for the clustering part, we coarsely follow 
Goemans and Williamson \cite{maxcutSDP}. 
For $W^{(1)}$ and $W^{(2)}$ being the intra-graph affinity matrices of graphs $\mathcal{G}_1$ and $\mathcal{G}_2$, respectively, the SDP relaxation reads
\begin{align} \label{eq:GC_relaxation} 
     &\underset{y^{(1)}, y^{(2)}}{\max} && \sum_{i,j} W^{(1)}_{ij}\Big(1-L^{(1)}_{ij}\Big) + \sum_{i,j} W^{(2)}_{ij}\Big(1-L^{(2)}_{ij}\Big) \\
     & \text{s.t.} && \begin{bmatrix} 
                        1 & \left(y^{(1)}\right)^T \\
                        y^{(1)} & L^{(1)}
                        \end{bmatrix} \succeq 0,
                        \begin{bmatrix}
                        1 & \left(y^{(2)}\right)^T \nonumber \\
                        y^{(1)} & L^{(2)}
                        \end{bmatrix} \succeq 0, \\
    & &&-1 \leq y^{(1)}_i \leq 1, -1 \leq y^{(2)}_i \leq 1, \nonumber \\
    & && L^{(1)}_{ii} = 1, \;\text{and}\;\, L^{(2)}_{ii} = 1. \nonumber 
\end{align} 
The SDP constraints (indicated by ${\succeq}\,0$) are convex relaxations of the non-convex quadratic equality constraint $L^{(\bullet)} = y^{(\bullet)} (y^{(\bullet)})^T$ (see,  \textit{e.g.,}~\cite{yenamandra2019convex}), under which the formulation in Problem~\eqref{eq:GC_relaxation} would solve the exact MAX CUT problem introduced in Sec.~\ref{ssec:joint_formulation}. 

\textbf{Joint Graph Matching and Clustering.} 
To obtain our joint graph matching and clustering formulation, we combine the convex relaxation of Problem~\eqref{eq:GM_relaxation}  with Problem~\eqref{eq:GC_relaxation}, and add a convex relaxation of the coupling constraint \eqref{eq:coupling_equation}. In the following, we explain the convex relaxation of the coupling constraint, for which we define the variable
\begin{equation}
        z:= \frac{1}{2}\begin{bmatrix}
    1 + y^{(1)} \\
    1 + y^{(2)}
    \end{bmatrix} \in [0,1]^{2n}
\end{equation}
that stacks the cluster vectors $y^{(1)}, y^{(2)}$ and maps the elements from $[-1,1]$ to $[0,1]$.
Under the assumption that $\overline{L} := zz^T$, the coupling constraint is equivalent to
\begin{align}\label{eq:coupling1}
    X_{ij} \leq \overline{L}_{i,j+n}, 
\end{align}
where the index offset $n$ is used to refer to the upper-right $n \times n$ block of the $2n \times 2n$ matrix $\overline{L}$.
However, since  $\overline{L} = zz^T$ is a non-convex constraint, analogously as in Problem~\eqref{eq:GC_relaxation}, it is relaxed to the convex constraint 
\begin{align}\label{eq:coupling2}
    \begin{bmatrix}
    1 & z^T \\
    z & \overline{L}
    \end{bmatrix} \succeq 0.
\end{align}
Overall, our joint graph matching and clustering problem is formed by simultaneously solving the convex relaxation of Problem~\eqref{eq:GM_relaxation} and Problem~\eqref{eq:GC_relaxation}, which are coupled via the constraints $\eqref{eq:coupling1}$ and $\eqref{eq:coupling2}$. The resulting problem is a convex semidefinite programming problem, which we solve using the off-the-shelf solver Mosek~\cite{MOSEK} interfaced through the modelling tool YALMIP~\cite{Lofberg2004}.
In order to obtain a permutation matrix from the continuously relaxed $X$, we follow the procedure in~\cite{PMSDP}. The continuously relaxed cluster variables $y^{(1)}$ and $y^{(2)}$ are converted to binary variables by  thresholding.
Algorithm \ref{Alg2} summarises the proposed joint  approach for graph matching and clustering. 
\begin{algorithm}[t!] 
    \caption{Joint Graph Matching and Clustering}
    \hspace*{\algorithmicindent} \textbf{Input:}$\;\;\;\,$affinity  matrix $K$, embbeded graphs $D,E$ \\ 
    \hspace*{\algorithmicindent} \textbf{Output:}  registration $X$, clustering $y_1, y_2$ 
    \begin{algorithmic}[1]
        \State calculate $W^{(1)}$ and $W^{(2)}$ as explained in Sec.~\ref{ssec:problem_formulations} 
        \State factorise  $\{ A_i \}_{i \in I}, \{ B_i \}_{i \in I}$ = $\operatorname{KPSVD}(K)$ 
        \For{$A_i \in \{ A_i\}_{i \in I}$}
        \State $P^{(1)}_{i} , P^{(2)}_{i}$ = $\operatorname{HOPE}(A_i)$ 
        \EndFor
        \For{$B_i \in \{ B_i\}_{i_I}$}
        \State $Q^{(1)}_{i} , Q^{(2)}_{i}$ = $\operatorname{HOPE}(B_i)$ 
        \EndFor
        \State $[R^{(1)}_{i}, R^{(2)}_{i}, X, y^{(1)},y^{(2)}]$ = \\
                $\operatorname{solveOurSDP}\big(\{P^{(1)}_{i} , P^{(2)}_{i} \}_{i \in I},\{Q^{(1)}_{i} , Q^{(2)}_{i} \}_{i \in I}\big)$ 
        \State binarise $X,y^{(1)},y^{(2)}$ 
        \State return $X,y^{(1)},y^{(2)}$ 
    \end{algorithmic} 
    \label{Alg2} 
\end{algorithm} 

\section{Experimental Evaluation}\label{sec:experimental_evaluation} 
\textbf{Datasets.} 
In our experiments, we first evaluate the graph matching accuracy of our method on the CMU House dataset \cite{CMUHouse}. 
Subsequently, we utilise synthetic data to test graph clustering and matching jointly, 
\begin{figure*}[t!] 
    \centering
    \includegraphics[width = \textwidth]{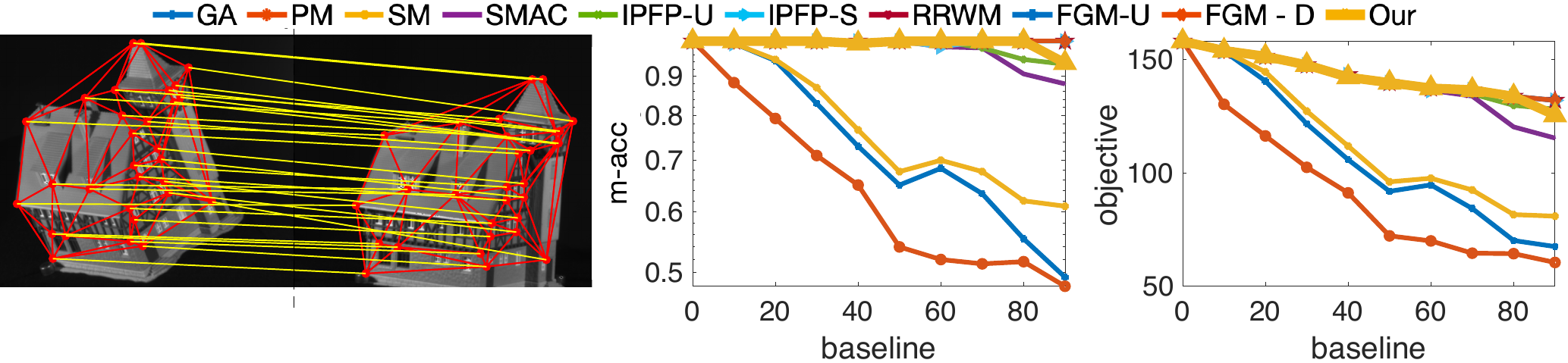}
    \vspace{-8mm}
    \caption{\textit{Left:} Example matching between frames $1$ and $90$. Yellow lines represent correct matchings. 
    \textit{Right: } Performance on the CMU House dataset using all $30$ nodes.}
    \label{fig:houseAcc}
\end{figure*}
where we particularly consider geometric primitives, such as prisms and pyramids, where the 
corners are selected as nodes, and the edges are obtained via Delaunay triangulation. 
In Sec.~\ref{ssec:qualitative_evaluation}, we provide qualitative results of our method on 3D scenes generated from the Princeton shape dataset \cite{princtonShape}. 
\textbf{Experimental Setup.} 
All algorithms, the benchmark from Fig.~\ref{fig:houseAcc}, SGM \cite{SeededGM} and ours, are implemented with Matlab \cite{MATLAB2020} 
(where some internal procedures of  FGM~\cite{fgm} are implemented in C++  for performance reasons). 
The experiments on the CMU House dataset are performed on a server with an EPYC 7702 CPU and 128GB RAM, and the tests with synthetic data are done on a workstation with an i7-9700 CPU and 32GB RAM. 

\subsection{Pure Graph Matching on CMU House \cite{CMUHouse}}\label{sssec:CMU_House}

We use the benchmark proposed along with the FGM algorithm \cite{fgm}. 
Here, $30$ landmarks are given on pairs of images, as can be seen in Fig.~\ref{fig:houseAcc}~(left). 
Delaunay triangulation is used to generate edges, and the Euclidean distance of those features is taken as the edge weight. We do not consider node affinities, \textit{i.e.,} the diagonal of $K$ is zero. 
As edge affinity for the edges with weights $e_{c_1}, e_{c_2}$, we use the   function
\begin{align*}
    \phi (e_{c_1}, e_{c_2}) = \text{exp} \left(- \frac{(e_{c_1} - e_{c_2})^2}{2500} \right). 
\end{align*}
In this setting we compare our algorithm against other state-of-the-art methods, \textit{i.e.,} 
Graduated Assignment (GA) \cite{ga}, Probabilistic Matching (PM) \cite{pm}, Spectral Matching (SM) \cite{sm}, Spectral Matching with Affine Constraints (SMAC) \cite{Cour2006}, Integer Projected Fixed Point method (IPFP) \cite{Leordeanu2009}, Re-weighted Random Walk Matching (RRWM) \cite{RRWM} and FGM \cite{fgm}. 
The accuracy of matchings is defined as:
\begin{align} 
    \operatorname{m-acc}= \frac{ \operatorname{tr}\left(X^\mathsf{T} X_{gt}\right)}{ \operatorname{tr}\left(\mathbf{1}_{n \times n} X_{gt}\right)}, 
    \label{eq:matchingAcc}
\end{align} 
where $X_{gt}$ is the ground-truth solution. 
As objective, we evaluate the Lawler form of the graph matching problem \eqref{eq:lawler}, which is shown in the right-most plot in Fig.~\ref{fig:houseAcc}. 
Our algorithm achieves high accuracy on the FGM benchmark. 
The plots in Fig.~\ref{fig:houseAcc} summarise the results 
and show the matching behaviour when all 30 features of the graphs are used. 
In this setting, we nearly always obtain a perfect matching and are comparable to the state-of-the-art FGM method on this dataset. 
We significantly outperform GA, PM, IPFP (two variants), SM and SMAC. 

\textbf{Timing and Parameter Selection.} We select $k =7$ for KPSVD in these experiments. Although our embedding dimensions $d$ fluctuate depending on the summands, we select them between $10$ and $27$. 
Using the Mosek solver \cite{MOSEK}, solving an instance takes ${\approx}1.5$h. Although our method is not competitive in terms of runtime in the pure graph matching setting, our SDP formulation competes with the state of the art in terms of accuracy, while also being highly flexible, as we demonstrate next.

\subsection{Matching and Clustering on Synthetic Data} 
For the joint evaluation of graph matching and clustering, we consider two metrics. 
The first one is the joint matching accuracy \textbf{m- \& c-acc}, which measures the joint graph matching and clustering performance of all algorithms, given by 
\begin{align} 
    \text{m- \& c-acc} := \sqrt[3]{\text{ m-acc } \cdot \text{  F-score $\mathcal{G}_1$ } \cdot \text{ F-score  $\mathcal{G}_2$}}. 
    \label{eq:totalAcc} 
\end{align} 
``F-score $\mathcal{G}_1$''---which is the clustering measure defined like in \cite{chicco2020advantages} and our supplement---denotes the clustering F-score achieved for the first graph $\mathcal{G}_1$. 
The second metric, \textbf{c-acc} considers the pure clustering performance, and is defined as
\begin{align}
 \text{c-acc} := \sqrt{\text{ F-score $\mathcal{G}_1$ }  \cdot \text{ F-score $\mathcal{G}_2$ }}. 
 \label{eq:clusterAcc}
\end{align}
\begin{figure}[t!]
    \centering 
    \includegraphics[width = 1.0 \linewidth]{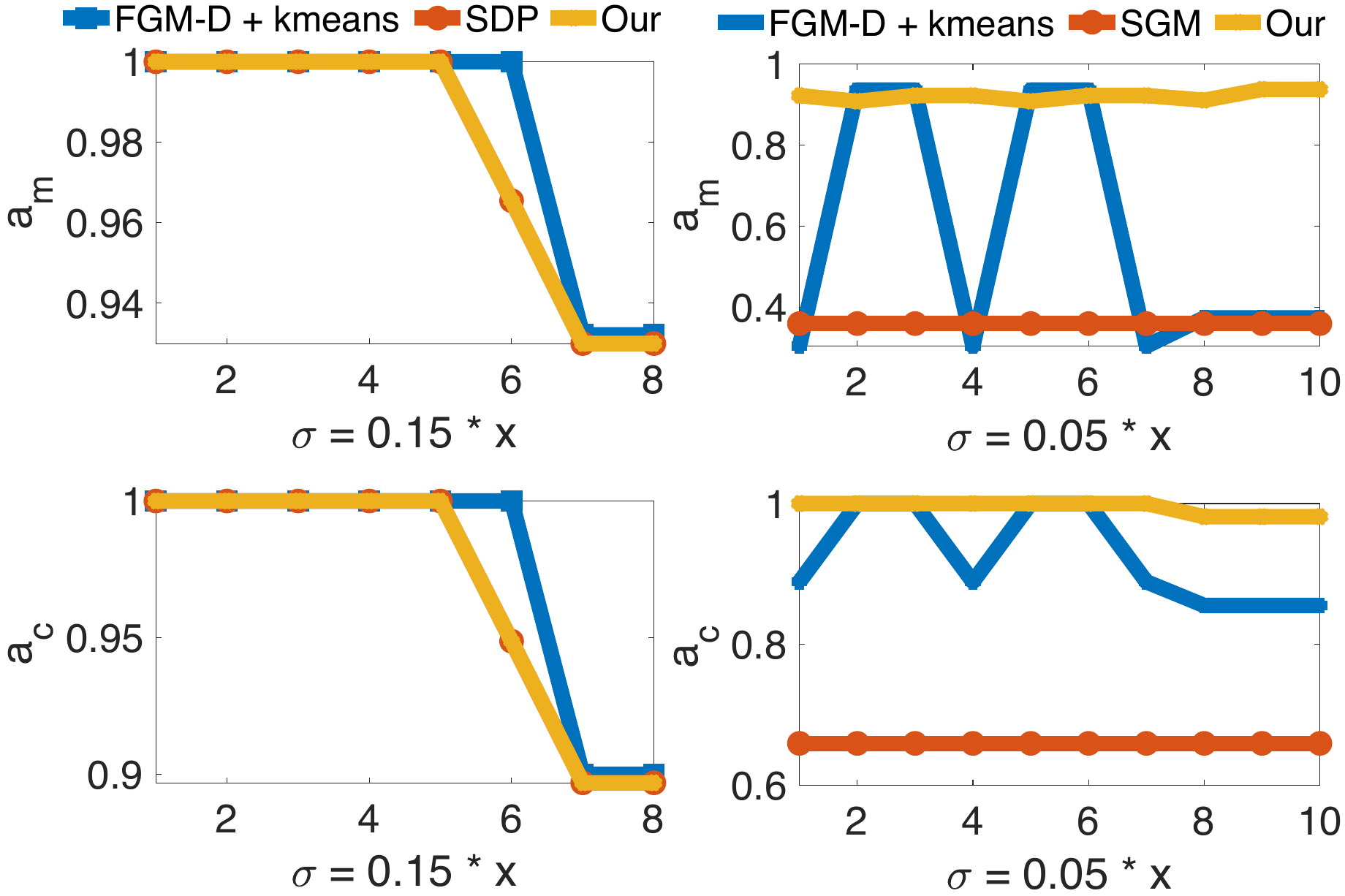} 
    \vspace{-7mm}
    \caption{ 
    Graph matching with increasing levels of Gaussian noise added to the node coordinates (left: $11$ nodes, right: $28$ nodes).
    \textit{Left:} 
    Our method performs as accurate as Schellewald's clustering-extended method, but ours is an order of magnitude faster. 
    \textit{Right:} Our method substantially  outperforms the clustering-extended FGM+D and SGM. 
    }
    \label{fig:1stgraphs} 
\end{figure} 

In the following, we demonstrate that jointly solving graph matching and clustering  enhances the accuracy of both tasks. 
Moreover, our experiments demonstrate that our more efficient joint relaxation is still as accurate as a joint relaxation based on the inefficient standard graph matching SDP relaxation \cite{schellewald}.
In order to compare against other methods, we extend existing two-graph matching algorithms for joint graph matching and clustering. 
For the SDP relaxation of Schellewald \textit{et al.} \cite{schellewald}, we do this by including our graph clustering SDP formulation  together with our coupling constraints~\eqref{eq:coupling_equation}.
The FGM-D method is extended na\"ively, \textit{i.e.,}~we assume that matched points should belong to the same cluster. 
After the matches are obtained, we perform $k$-means clustering on the embedded nodes (the coordinates are available from the data). 
We select FGM-D as graph matching method, due to its superior performance on the CMU House dataset (see Sec.~\ref{sssec:CMU_House}).

\textbf{Increasing Noise Magnitudes.} 
We perform experiments on noisy data with increasing magnitude of Gaussian noise (with standard deviation $\sigma$) applied to the point coordinates of the original data. 
In Fig.~\ref{fig:1stgraphs}-(left column), we demonstrate that for smaller graphs ($11$ nodes), our relaxation delivers equally accurate results compared to Schellewald's matching with the coupled clustering constraints. 
At the same time, our method has the accuracy comparable to FGM-D.
This small graph setting serves as a sanity check that our  relaxation is as accurate as Schellewald \textit{et al.}~\cite{schellewald}. 
Further, we consider graphs with $28$ nodes, see 
Fig.~\ref{fig:1stgraphs}-(right). 
The accuracy of our method is virtually not affected by the noise, whereas the performance of FGM-D+$k$-means fluctuates substantially. 
We also compare against SGM \cite{SeededGM}, which addresses both problems in sequence (clustering first, then graph matching). 
We observe that SGM can neither compete with the clustering-extended FGM-D, nor with our method. 
Note that due to runtime constraints, we do not include the variant of Schellewald's algorithm with clustering in this and the next experiments. 

\begin{figure} 
    \centering 
    \includegraphics[width =  \linewidth]{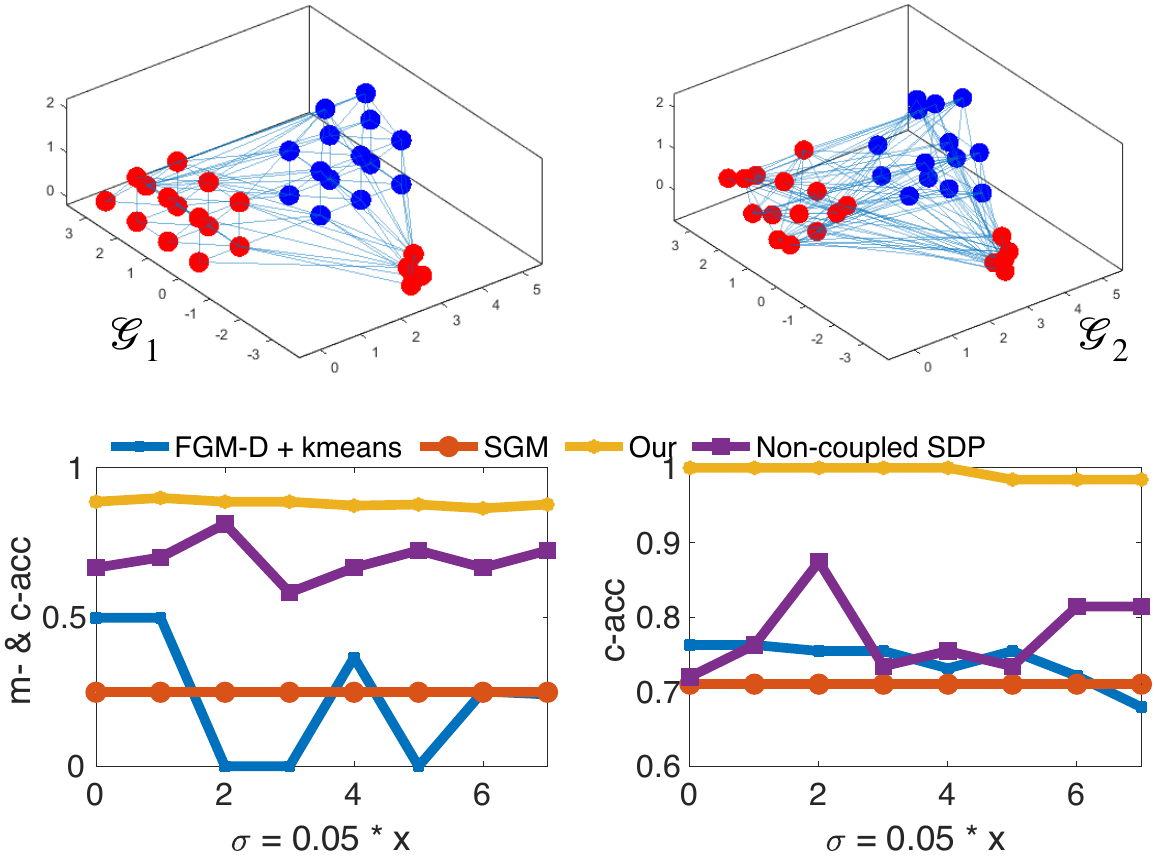} 
    \caption{\textit{Top Row:} Qualitative results on synthetic data with Gaussian noise and non-uniform clusters ($33$ nodes). 
    \textit{Bottom Row:} Quantitative results. Our coupling constraint increases the matching and clustering accuracy. 
    } 
    \label{fig:clusteredOutlier} 
\end{figure} 

\textbf{Non-Uniform Clusters.} 
Challenging scenarios, both for clustering and matching algorithms, are those with non-uniform clusters, \textit{i.e.,} those which contain outliers with respect to the predominant cluster points. 
To test how different methods perform on such data, we add clustered outliers to the samples with increasing levels of Gaussian noise. 
Such---at first glance---insignificant change of both graphs, appears too challenging for the competing methods FGM+D and SGM. 
\begin{figure*}[t!]
    \centering 
    \includegraphics[width = \linewidth]{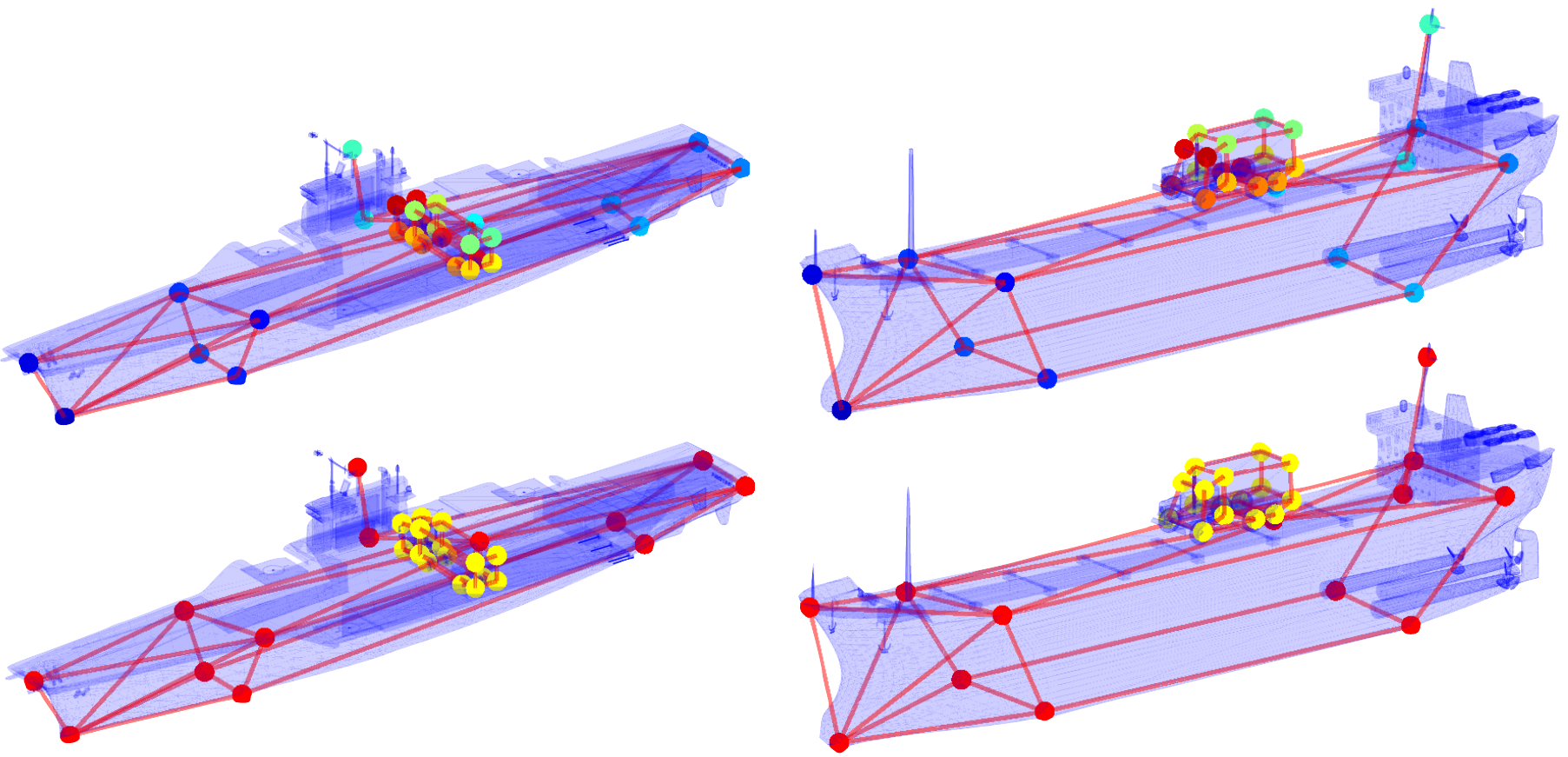} 
    \caption{Matching and clustering of a shape pair from the Princeton shape dataset. 
    \textit{Top row:} Our algorithm correctly matches the nodes (the matches are colour-coded). 
    \textit{Bottom row:} Identified clusters (red and yellow points). 
    } 
    \label{fig:combinedGraphic} 
\end{figure*} 
Here, we force the clustered outliers to be always assigned to the same cluster. 
Results are shown in Fig.~\ref{fig:clusteredOutlier}, where we can observe that our joint approach outperforms existing algorithms in joint clustering and matching performance, as well as in terms of the pure clustering evaluation. 
Also notable is that our method without the clustering constraint outperforms FGM and SGM, but is worse than our relaxation including the clustering constraint. 

\textbf{Timing and Parameter Selection.} 
For the experiments with $11$ nodes, we choose $k=6$ in the KPSVD. Our embedding dimensions $d$ are either $2$ or $3$.
It took a total of $4$ seconds to solve this problem with our SDP  relaxation, whereas Schellewald's method with the clustering extension takes ${\approx}50$ sec. 
For the experiments with $28$ nodes, we choose $k=7$ in the KPSVD. The embedding dimension $d$ is set between $6$ and $9$.
In this case, solving the problem took Mosek~\cite{MOSEK} $150$ seconds (pre- and  post-processing takes around one second). 

\subsection{Qualitative Evaluation with 3D  Scenes}\label{ssec:qualitative_evaluation} 

We next show practical applications of joint graph matching and clustering. 
We choose several shapes from the Princeton shape dataset \cite{princtonShape}, from which we construct medium-sized graphs with distinct high-level features. 
Thus, two helicopters and two cylindrically-symmetric vases allow for a meaningful matching even though the shapes of the selected samples differ, see Fig.~\ref{fig:teaserHeli}. 
The clustering correctly identifies the high-level objects vase and helicopter. 
In the example visualised in Fig.~\ref{fig:combinedGraphic}, we also see how our method correctly matches the nodes of an aircraft carrier and a cargo ship, while at the same time
correctly identifying two clusters, one of which being the ships, and the other one being trucks on top of the deck of the ships, each of them being differently oriented in the input data. 
In these experiments, we take the first five summands of the Kronecker decomposition, which results in $90\%$ of the information. 
Each run takes $25$ sec on i7-9700 CPU with mosek \cite{MOSEK}. 
Pre- and post-processing requires one second. 
\section{Discussion} 
Our experiments confirm that the proposed algorithm efficiently solves graph matching and clustering, thanks to the joint formulation.
In preliminary tests, we observed that our method could have a lower accuracy for non-bijective settings. 
We believe that this is inherited from the SDP relaxation of PMSDP \cite{PMSDP} since the convex relaxation is less tight in such cases. 
At the moment, our method is tailored towards two clusters, and 
extending it towards multiple clusters will require a different SDP relaxation. 
The accuracy of our method in scenarios with noise---when matching nodes are perturbed---is higher compared to several other methods that we evaluated. Naturally, due to the spectral nature of the HOPE embedding \cite{HOPE}, our method is not entirely agnostic to noise.  
Even though we tested about a dozen different graph embeddings, including~\cite{spectralEmbedding, GLEE, HOPE, deepwalk, tsitsulin2018verse, rozemberczki2019gemsec}, we cannot preclude that our technique can be further improved with another more effective graph embedding in future. 
The computational efficiency of our approach depends on the SDP solver. 
Any progress in terms of the efficiency of SDP solvers  will reduce the runtime of our method. 

\section{Conclusion} 

We present a new method for simultaneous graph matching and clustering, which does not depend on the availability of labelled training data. 
Thanks to our novel convex SDP relaxation with coupling  constraints, we can guarantee solution optimality of the relaxed objective. 
Experiments confirm that our method 
achieves higher accuracy on average across multiple evaluation scenarios,  and more consistent results overall, compared to the current state of the art. 
We believe that our proposed technique offers multiple new avenues for future research. 

{\small
\bibliographystyle{ieee_fullname.bst}
\bibliography{egbib}
}

\onecolumn
\setcounter{section}{0}
\renewcommand\thesection{\Alph{section}}
\newcommand{\suppsection}{\subsection}
\clearpage
\begin{center}
\textbf{\Large Supplementary Material\vspace{10pt}} 
\end{center} 
\makeatletter

This supplementary material is divided into two parts. 
The definition of the HOPE embedding according to Ou \textit{et al.}~\cite{HOPE} can be found in Sec.~\ref{sec:HOPE_embedding}. 
In addition, in Sec.~\ref{sec:F_score}, the definition of the F-score is  given, which we use in the experimental section of the main paper (Sec.~\ref{sec:experimental_evaluation}). 

\appendix

\section{HOPE Embedding}\label{sec:HOPE_embedding} 
HOPE embedding takes a similarity measure matrix $S$.
A similarity measurement measures the similarity between nodes, similar to the intra-graph affinity matrix $W$ defined in Sec.~\ref{ssec:problem_formulations}. 
Our experiments indicate that the measurement of \textit{common neighbours} introduced in the HOPE paper \cite{HOPE} is the most noise-resistant measurement. 
It is defined such that $S:= A^2$, where $A$ is the graph's adjacency matrix. 
The embedding method calculates two sets of embedding vectors $P$ and $Q$, by performing  $\operatorname{svd}$ of the similarity matrix $S$: 
\begin{equation} \label{hope}
    \begin{split}
    P &= \big[\sqrt{\sigma_1} p_1, \dots, \sqrt{\sigma_d} p_d\big], \,\text{and} \\ 
    Q &= \big[\sqrt{\sigma_1} q_1, \dots, \sqrt{\sigma_d} q_d\big]. 
    \end{split}
\end{equation}
where $\{\sigma_1, \dots, \sigma_d \}$ are the $d$ largest singular values of $S$ arranged in descending order. 
HOPE embedding algorithm then minimises the following loss function: 
\begin{align}\label{eq:HOPE_loss} 
    \mathcal{L}(P, Q) = \norm{S - P {Q}^\mathsf{T}}^2_F, 
\end{align} 
where $\norm{\cdot}_F$ denotes the Frobenius norm. 

\vspace{4pt} 

\section{F-Score}\label{sec:F_score} 

The F-score is defined as \cite{chicco2020advantages}: 
\begin{align}
    F = \frac{TP}{TP + \frac{1}{2}(FP + FN)}. 
\end{align}
Here, TP (\textit{true positives}) is the number of cases when nodes of a node pair are correctly assigned as belonging to a given cluster. 

FP (\textit{false positives}) is the number of cases when nodes from a node pair are assigned to the same cluster, whereas they belong to different ones.
FN (\textit{false negatives}) is the number of cases when nodes from a node pair are not assigned to the same cluster, whereas they belong to a single one. 

\end{document}